# Comparing Efficiency of Expert Data Aggregation Methods

Sergii Kadenko[1][0000-0001-7191-5636] and Vitaliy Tsyganok[1][0000-0002-0821-4877]

[1] Institute for Information Recording of the National Academy of Sciences of Ukraine

**Abstract.** Expert estimation of objects takes place when there are no benchmark values of object weights, but these weights still have to be defined. That is why it is problematic to define the efficiency of expert estimation methods. We propose to define efficiency of such methods based on stability of their results under perturbations of input data. We compare two modifications of combinatorial method of expert data aggregation (spanning tree enumeration). Using the example of these two methods, we illustrate two approaches to efficiency evaluation. The first approach is based on usage of real data, obtained through estimation of a set of model objects by a group of experts. The second approach is based on simulation of the whole expert examination cycle (including expert estimates). During evaluation of efficiency of the two listed modifications of combinatorial expert data aggregation method the simulation-based approach proved more robust and credible. Our experimental study confirms that if weights of spanning trees are taken into consideration, the results of combinatorial data aggregation method become more stable. So, weighted spanning tree enumeration method has an advantage over non-weighted method (and, consequently, over logarithmic least squares and row geometric mean methods).

## 1 Introduction

Expert estimation is a powerful decision support tool for weakly structured subject domains. Weakly structured subject domain features are listed in many sources, for instance, in [1]. In our current research we propose to focus on such features of weakly structured domains as lack of benchmarks, incompleteness of information on estimated objects, and impact of human factor. These features make it difficult to define the efficiency of expert estimation methods.

While measurement of objects according to quantitative parameters (such as length, weight, duration etc), actually, means comparing them to some benchmark values or units (foot, pound, second etc), people resort to expert estimation as an alternative to measurement in cases when measurement of objects is impossible. In such cases expert estimates of objects become the only source of quantitative information about these objects. Experts can provide both direct estimates in certain scales (ordinal or cardinal, numeric or verbal, agreement scale etc) and pair-wise comparisons of objects. According to many specialists, the best way to measure a set of objects according to some "intangible" criterion is to compare them with each other. This assumption resulted in emergence of many methods based on pair-wise comparisons of objects. Particularly, we should mention the Analytic Hierarchy/Network Process (AHP/ANP) [2, 3], TOPSIS [4], "triangle" and "square" [5, 6], combinatorial method [5-7], logarithmic least squares method (LLSM) [8].

Efficiency indicators for methods which operate with determined data are mostly based on different measures of deviation of real (experimental) data from benchmark values (average mean (square) deviation, mathematical expectation of error, Euclidean distance etc). When it comes to expert data-based methods, there is always a question: "what should we compare expert data (and results of their processing) with?" (as, again, there are no benchmarks). When a decision-maker (DM) organizes an examination, (s)he heuristically assumes that there is some ground truth, i.e. "exact" values of estimates of objects and ratios between them. So, the key question is: which indicators can define the degree of accuracy and credibility of expert data and methods of their aggregation? Level of DM trust towards expert recommendations and decisions, made on their basis, will depend on these indicators.

Academic publications list relatively few approaches to determination of accuracy of expert methods (if the term "accuracy" applies at all). According to the academic school of T.Saaty, the main requirement to expert data, input into pair-wise comparison matrices (PCM), is ordinal and cardinal consistency (absence of transitivity violations) [2]. In [3] the term "legitimization" is used. Expert data-based examination result is "legitimate" if it coincides with the DM's independent choice (in [3] there is a curious example of location selection for Disneyland in China). Pankratova and Nedashkovskaya [9] demonstrate that results, obtained using ANP and original hybrid method, coincide, and this, according to the authors, confirms the credibility of hybrid method. In Elliot's research [10] experts compare several estimation scales and define which of these scales allows them to express their preferences in the most adequate way. A similar approach is used in [11], where the experts choose the result of aggregation of their estimates, which reflects their understanding of the subject domain most adequately.

The common feature of the listed approaches is absence of any benchmark estimate values (in actual expert examinations there are, indeed, no benchmarks). Conceptually different approach involves testing of expert methods on the specially generated (simulated) set of "benchmark" (model) objects, for which the exact values of their estimates according to a certain criterion are known. For example, we can mention an experiment [2], where respondents are asked to estimate the ratios of several figure squares. Exact ratios are known only to the experiment organizer. These model values are compared with values, obtained based on expert estimates using group AHP.

In Ukraine a similar approach is used in [5, 6]. The authors compare around 20 expert methods according to 3 criteria: accuracy (precision), duration of estimation process, and consistency of its results. Experts estimate 7 objects (colored figures) using different methods. Number 7 is chosen due to psycho-physiological limitations of human mind [12]. Real (benchmark, model) ratios between object weights are, again, known only to the experiment organizer. Aggregate estimation results are compared to benchmark values. Methods that produce smaller average error are considered more accurate.

Still another approach to defining accuracy of expert methods is based on simulation and does not require participation of experts at all. Both model object weights and expert estimates are simulated. Such simulation of PCM is used to define threshold values of consistency index (CI) and ratio (CR) (these values are constantly updated based on the number of modeled PCM [3]). Examples of PCM simulations can also be found in [13].

## 2   Combinatorial method: an overview

For the first time combinatorial method of pair-wise comparison aggregation was introduced in the early 2000-s; since then it went through many updates and improvements [7]. The problem (just like in AHP) is to find a vector of $n$ priorities (object weights) based on PCM, provided by one or several experts. The key idea of the method is most thorough usage of expert data, provided in the form of PCM. In the general case, this information is redundant. That is why, all non-redundant informatively-meaningful basic pair-wise comparison sets, which can be formed from elements of a given PCM, are enumerated. Sometimes these basic sets are called spanning trees (the term is borrowed from graph theory). According to Cayley's theorem on trees [14], a complete PCM with dimensionality $n \times n$ allows us to form $n^{n-2}$ of such spanning trees. Each basic set allows us to build a vector of relative object weights. After that we can find the aggregate priority vector as ordinary (1) or weighted (2) average.

Practical implication of combinatorial method is its usage as the primary expert estimate aggregation tool in the strategic planning technology for weakly-structured subject domains [11].

In 2010 the advantage of combinatorial method over other pair-wise comparison aggregation methods was demonstrated [15]. In 2012 the method was "re-invented" [16, 17]. In 2017 updated formulas for calculation of relative expert competence coefficients (based on the quality of expert information) were introduced [18]. During the last few years equivalence of combinatorial method, LLSM [8] (for complete and incomplete, additive and multiplicative PCM), and row geometric mean [19] methods was proved. However, equivalence holds only if ordinary (and not weighted) average formula is used for aggregation (1).

$$w_j^{aggregate} = (\prod_{q=1}^{T} w_j^q)^{1/T} ; j = 1..n \qquad (1)$$

where $T \leq mn^{n-2}$ is the total number of basic pair-wise comparison sets (spanning trees); $\{w_j^q; j=1..n; q=1..T\}$ is the set of relative weights of $n$ objects, calculat-ed based on spanning tree number $q$; ($w_j^{aggregate}, j=1..n$) are aggregate object weights; $m$ is the number of experts.

Conceptual difference of the modified combinatory method is usage of ratings of basic pair-wise comparison sets. These ratings are based on *completeness, detail, consistency,* and *compatibility* of data, input by experts into individual PCM. Ratings of priority (relative alternative weight) vectors, obtained from ideally consistent PCM, reconstructed based on every single spanning tree (basic pair-wise comparison set), are taken into consideration. As a result, weight aggregation formula (1) assumes the following look (2):

$$w_j^{aggregate} = \prod_{k,l=1}^{m} (\prod_{q^k=1}^{T_k} (w_j^{(kq_k l)})^{\frac{R_{kq_k l}}{\sum_{u,p,v} R_{upv}}}); j=1..n \quad (2)$$

Additive look of basic pair-wise comparison set (spanning tree) rating is as follows:

$$R_{kql} = c_k c_l s^{kq} s^l \Big/ \ln(\sum_{u,v} |a_{uv}^{kq} - a_{uv}^{l}| + e) \quad (3)$$

Multiplicative look of basic pair-wise comparison set (spanning tree) rating is as follows:

$$R_{kql} = c_k c_l s^{kq} s^l \Big/ \ln(\prod_{u,v} \max(\frac{a_{uv}^{kq}}{a_{uv}^{l}}; \frac{a_{uv}^{l}}{a_{uv}^{kq}}) + e - 1) \quad (4)$$

In formulas (3) and (4) $k, l$ are the numbers of experts ($k, l = 1..m$), whose PCM are being compared with each other; $c_k, c_l$ are a-priori values of relative expert competence; $k$ and $l$ can be equal or different; $q$ is the number of ideally consistent PCM copy $q = 1..mT_k$; $s^{kq}$ is the relative average weight of scales in which basic pair-wise comparison set elements are input; it is calculated based on Hartley's formula [20];

$$s^{kq} = (\prod_{u=1}^{n-1} \log_2 N_u^{(kq)})^{\frac{1}{n-1}} \quad (5); \quad s^k = (\prod_{\substack{u,v=1 \\ v>u}}^{n} \log_2 N_{uv}^{(k)})^{\frac{2}{n(n-1)}} \quad (6)$$

$s^l$ is the average weight of scales, in which elements of the respective individual PCM of expert number $l$ are input.

In (5) and (6) $N$ is the number of grades in the scale, in which the respective pair-wise comparison is provided. More detailed explanation of spanning tree ratings can be found in [18].

## 3 Numeric example

3 equally competent experts $E_1, E_2, E_3$ ($c_1 = c_2 = c_3 = 1$) compare 4 objects ($A_1, A_2, A_3, A_4$) in integer scales. We should calculate relative object weights (priorities) based on PCM, provided by the experts. Total numbers of grades in the scales, selected by experts, are given in Table 1. Table 2 provides particular grade numbers, selected by the experts. Table 3 provides the PCM, brought to the unified scale.

**Table 1**. Number of grades in the scales, selected by experts for pair-wise comparisons

|       | $E_1$ |       |       |       | $E_2$ |       |       |       | $E_3$ |       |       |       |
|-------|-------|-------|-------|-------|-------|-------|-------|-------|-------|-------|-------|-------|
|       | $A_1$ | $A_2$ | $A_3$ | $A_4$ | $A_1$ | $A_2$ | $A_3$ | $A_4$ | $A_1$ | $A_2$ | $A_3$ | $A_4$ |
| $A_1$ | 1 | 9 | 8 | 7 | 1 | 3 | 4 | 5 | 1 | 9 | 9 | 8 |
| $A_2$ |   | 1 | 6 | 5 |   | 1 | 6 | 7 |   | 1 | 3 | 9 |
| $A_3$ |   |   | 1 | 4 |   |   | 1 | 8 |   |   | 1 | 7 |
| $A_4$ |   |   |   | 1 |   |   |   | 1 |   |   |   | 1 |

**Table 2**. Numbers of specific grades of pair-wise comparison scales, selected by the experts

|       | $E_1$ |       |       |       | $E_2$ |       |       |       | $E_3$ |       |       |       |
|-------|-------|-------|-------|-------|-------|-------|-------|-------|-------|-------|-------|-------|
|       | $A_1$ | $A_2$ | $A_3$ | $A_4$ | $A_1$ | $A_2$ | $A_3$ | $A_4$ | $A_1$ | $A_2$ | $A_3$ | $A_4$ |
| $A_1$ | 1 | 2 | 4 | 7 | 1 | 3 | 4 | 5 | 1 | 2 | 4 | 8 |
| $A_2$ |   | 1 | 2 | 4 |   | 1 | 2 | 3 |   | 1 | 2 | 5 |
| $A_3$ |   |   | 1 | 2 |   |   | 1 | 2 |   |   | 1 | 3 |
| $A_4$ |   |   |   | 1 |   |   |   | 1 |   |   |   | 1 |

**Table 3**. Values of pair-wise comparisons, brought to the unified scale

|       | $E_1$ |       |       |       | $E_2$ |       |       |       | $E_3$ |       |       |       |
|-------|-------|-------|-------|-------|-------|-------|-------|-------|-------|-------|-------|-------|
|       | $A_1$ | $A_2$ | $A_3$ | $A_4$ | $A_1$ | $A_2$ | $A_3$ | $A_4$ | $A_1$ | $A_2$ | $A_3$ | $A_4$ |
| $A_1$ | 1 | 2 | 4 1/3 | 8 5/6 | 1 | 7 1/2 | 8 1/6 | 8 1/2 | 1 | 2 | 4 | 9 |
| $A_2$ | 1/2 | 1 | 2 2/7 | 6 1/2 | 1/7 | 1 | 2 2/7 | 3 1/2 | 1/2 | 1 | 3 1/2 | 5 |
| $A_3$ | 2/9 | 3/7 | 1 | 2 5/6 | 1/8 | 3/7 | 1 | 2 | 1/4 | 2/7 | 1 | 3 1/2 |
| $A_4$ | 1/9 | 1/6 | 1/3 | 1 | 1/8 | 2/7 | 1/2 | 1 | 1/9 | 1/5 | 2/7 | 1 |

48 ideally consistent PCM (ICPCM) ($mn^{n-2} = 3 \times 4^2 = 48$) are built based on initial 3 PCM. Each ICPCM is constructed from the respective basic pair-wise comparison set (spanning tree). For instance, the basic set of pair-wise comparisons of objects $(A_1, A_2)$, $(A_1, A_3)$, and $(A_2, A_4)$ corresponds to the spanning tree, shown on Fig. 1 (clockwise). From the respective elements of the PCM provided by the ex-pert $E_2$ we reconstruct the ICPCM, shown in Table 4 (basic pair-wise comparison values are highlighted in bold, while other elements are reconstructed based on transitivity rule).

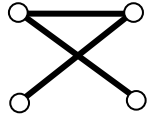

**Fig 1**. Spanning tree example for 4 objects

**Table 4**. ICPCM example

| 1 | **7 1/2** | **8 1/6** | 26 ¼ |
|---|---|---|---|
| 1/7 | 1 | 1 | **3 ½** |
| 1/8 | 1 | 1 | 3 1/5 |
| 0 | 2/7 | 1/3 | 1 |

Similarly, ICPCM are reconstructed from all basic pair-wise comparison sets, provided by each of the 3 experts. In this process, in order to verify consistency and compatibility of the initial PCM, ICPCM are compared with these initial PCM provided by the experts. For this purpose, 3 copies of each ICPCM are built. So, the total number of ICPCM to be analyzed is $T = m^2 n^{n-2} = 144$. Each ICPCM is assigned a rating, calculated according to (4). For, instance, when ICPCM, shown in table 4, is compared to the PCM of the first expert $E_1$, non-normalized value of the respective rating equals 1.191. When all ICPCM ratings are calculated, they are normalized by sum (see power index in (2)). From each ICPCM copy number $q = 1..144$ we reconstruct a vector of relative object weights $w_1^q, w_2^q, w_3^q, w_4^q$ (ideal consistency of the matrix allows us to use any basic set of pair-wise comparisons as priority vector; for instance, the first row). Finally, the aggregate priority (object weight) vector is calculated according to (2).

Normalized priorities $w_1, w_2, w_3, w_4$, calculated using the modified combinatorial method (2), based on the example data equal (0.563734299; 0.263382041; 0.120820159; 0.052063501). Values of priorities, calculated using ordinary combinatorial method (1) equal (0.590174795; 0.243658012; 0.114086692; 0.052080501).

It has been proven that ordinary combinatorial priority aggregation method (1) is equivalent to row geometric mean [19] and LLSM [8]. At the same time, as we can see from the example (and from [18]), results produced by ordinary (1) and modified method (2) are significantly different, so these two methods are not equivalent.

Both row geometric mean and LLSM have lower computational complexity than combinatorial method, and this is their advantage. The key advantage of modified combinatorial method is that it allows us to consider the quality of expert data prior to its aggregation. Consequently, results of its work more adequately reflect the level of expert competence in the issue under consideration.

The current research is an attempt to empirically confirm the advantage of the modified combinatorial method over the ordinary method (and, consequently, over row geometric mean and LLSM).

## 4    Available approaches to determination of efficiency of ordinary and modified combinatorial method

At the beginning of the paper we outlined several approaches used to determine the efficiency (and compare) expert methods. However, not all these approaches are applicable to our particular case.

1. Holding real expert sessions (Saaty's "legitimization" [3]) intended to empirically verify certain hypotheses is a "luxury" that an average researcher cannot afford. Finding real experts and obtaining estimates from them requires too many resources.

2. Comparing results of several methods [9] and expecting them to coincide is not our task under the circumstances. We are trying to define, which of the two methods produced *better* results.

3. Holding model examinations (for example, with students), such as ones described in [10, 11], where experts themselves would define, which results more adequately reflect their understanding of the subject domain, again would not solve the problem. We are trying to define objective characteristics of the methods that do not depend on the attitudes of the respondents, and to compare methods according to these characteristics.

4. Testing of the methods on the data of real expert estimation of specially modeled objects [5, 6] is plausible.

5. Simulation of model object weights and of expert estimates of ratios between them (priorities) [15] is plausible.

So, we propose to focus on the last two of the listed approaches: a) comparing of the two methods on real expert estimates of a set of model objects and b) simulation of both model object weights and expert estimates.

## 5    Comparing ordinary and modified combinatorial methods on real data of expert estimation of model objects

The model objects were figures with different numbers of colored pixels (known to expert session organizer). The number of such figures, which the experts compared according to coloring degree, amounted to 7 ($n = 7$) (based on psycho-physiological constraints of human mind [12]). 18 independent pair-wise comparison sessions were conducted with real respondents (experts). Pair-wise comparisons were multiplicative ones, that is PCM transitivity (consistency) requirement looked as follows: $a_{ij} = w_i/w_j = (w_i/w_k) \times (w_k/w_j) = a_{ik} \times a_{kj}; i, j, k = 1..n$, where $a_{ij}$ is the value of pair-wise comparison of objects number $i$ and $j$; $w_i, w_j, w_k$ are the weights of objects with respective numbers. Based on expert PCM, 4 series of calculations were performed, respectively, for individual and group, ordinary (1) and modified (2) combinatorial methods. Estimates were input in a unified scale and experts were considered equally competent a priori, so the numerator in the multiplicative rating formula (4) equaled 1.

Obtained weights were compared with true values and relative estimation errors were calculated for each of the 7 objects (7):

$$\delta_k = \frac{|w_k - w_k^{true}|}{w_k^{true}}; k = 1..n \qquad (7)$$

Group estimation sessions were simulated as combinations of groups of 3 experts from 18 available individual estimation precedents. So, the number of such group sessions amounted to $C_{18}^3 = 816$. The generalized accuracy indicator was the average relative estimation error calculated across all objects:

$$\delta = \frac{1}{n}\sum_{k=1}^{n} \delta_k = \frac{1}{n}\sum_{k=1}^{n} \frac{|w_k - w_k^{true}|}{w_k^{true}} \qquad (8)$$

In order to conduct the experiment we used the original software module (Fig. 2).

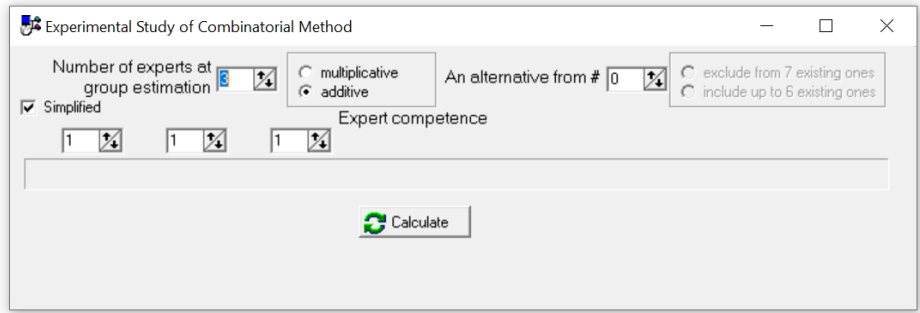

**Fig. 2.** Screenshot of the software module for experimental study of combinatorial method on real estimation data

Brief algorithm of the experiment (once the module is launched) is as follows.

1. Pair-wise comparison type (additive or multiplicative), as well as the number of experts in the group are selected, and their relative competence values are set. Combinatorial method modification (ordinary (1) or modified (2)) is determined. There is an opportunity to exclude some objects from the initial set of 7 model objects.

2. The module reads pair-wise comparison values from the file and performs calculations. It calculates priorities based on each estimation precedent using the selected combinatorial method modification. Calculation results (values of 7 model object weights) are written into another file in real time.

3. The module ends its work when all possible expert group variants are enumerated. If the general number of individual expert estimation precedents equals 18, and the number of experts in a group equals 3, then calculations are performed for $C_{18}^3 = 816$ group estimation sessions (1st session includes 1st, 2nd, and 3rd precedent; 2nd session – 1st, 2nd, and 4th precedent; ... 816th session – 16th, 17th, and 18th precedent).

Calculated individual estimation error values for modified ("weighted") and ordinary (simplified) methods are shown on Fig. 3. Group estimation errors are shown on Fig. 4. On both figures X-axis denominates numbers of expert estimation sessions of 7 model objects. Y-axis denominates average mean estimation errors (8) (values range from 0 to 1). Each number of estimation session is associated with 2 average relative error values (1 for ordinary and 1 for modified method), which correspond to 2 points on coordinate plane. If for some specific estimation session, the point, corresponding to one of the two methods, lies higher (has larger Y), it means that this method produces larger error (its results are worse in comparison to true value) on the respective set of expert estimates.

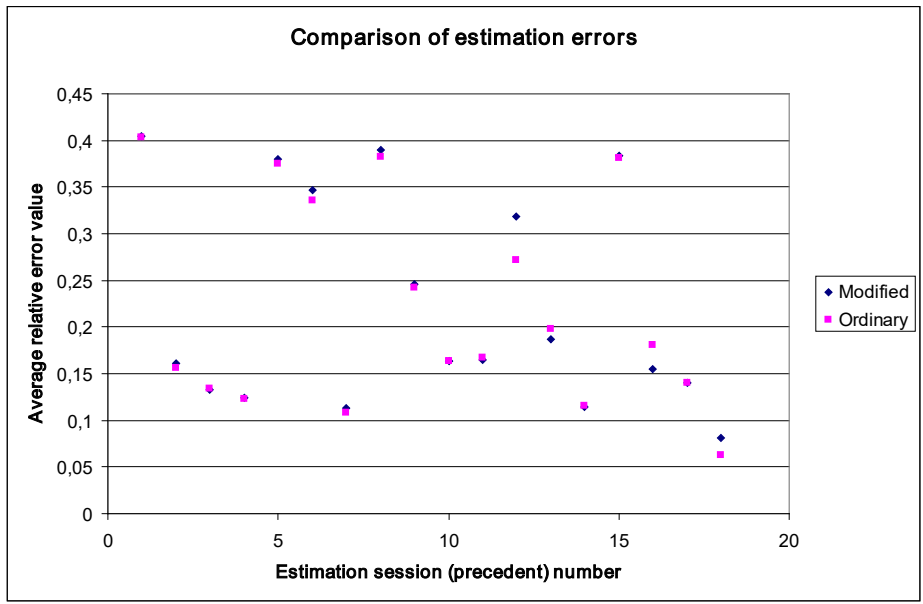

**Fig. 3**. Errors of estimation of 7 model objects using ordinary and modified combinatorial method (18 estimation sessions)

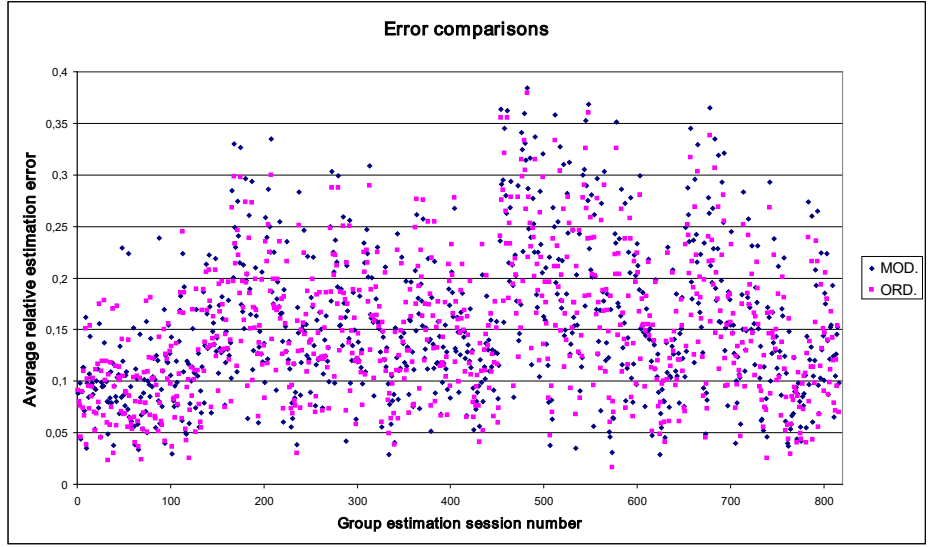

**Fig. 4**. Errors of estimation of 7 model objects using ordinary and modified group combinatorial method (816 estimation sessions)

Ranges and average values of errors (8) for 816 group estimation precedents are shown in table 5.

**Table 5**. Characteristics of modified and ordinary combinatorial methods based on real group expert estimation data

|  | Maximum average error | Minimum average error | Average error across all estimation precedents |
|---|---|---|---|
| Ordinary method | 0.379730445 | 0.016795423 | 0.126609639 |
| Modified method | 0.384422003 | 0.028881538 | 0.130089121 |

Results of experimental research of ordinary and modified combinatorial method on the data of real expert estimation of model objects do not allow us to draw any definite conclusion as to advantage of one of the two methods. As we can see from figures 3 and 4, on some individual and group estimation precedents, ordinary method (that does not take the weights of basic spanning trees into account) turns out to be more accurate, while on others – the "weighted" method yields more accurate results.

The look of priority aggregation formulas (1) and (2), as well as data from 816 group estimation precedents indicate, that the modified method tends to be more efficient in the cases, when the number of accurate comparisons (closer to model true values) exceeds the number of inaccurate ones. If the majority of comparisons is inaccurate (far from true values), although consistent and compatible, they "pull" the aggregate estimate value towards themselves, as a result of weighting procedure (2). Consequently, on such precedents, the ordinary method produces better results. Moreover, both methods can produce paradoxes when averaging of inconsistent values far from true ones still produces rather accurate aggregate result.

Intermediate conclusion: it is not the chosen estimate aggregation procedure, but the estimate values themselves, that influence the results of a method. During estimation session, expert's mind "constructs" its own model priority vector. Result and relative accuracy of the method's work depend on the proximity of the expert's assumptions to the actual true vector. Hence, the difference between priorities obtained through aggregation of real expert data, and true priority values (model object weights) cannot serve as an indicator of efficiency of aggregation methods, because accurate result of a method's work confirms only the accuracy of initial expert data.

It makes sense to use relative accuracy of real expert estimates of model objects as efficiency criterion for comparing conceptually different methods (for example, methods using multiplicative and additive scales; verbal, graphic, or numeric data input; pair-wise comparison vectors, triangular, or square PCM; methods with or without feedback etc), as shown in [5, 6]. When such methods are compared, input data and aggregation procedures are substantially different. Within our current research we use one and the same expert estimate set and similar aggregation procedures. That is why usage of the approach described in [5, 6] in the context of the present research turns out to be incorrect, and produces unrepresentative results.

Human factor (subjective notion of the expert regarding ratios between objects) adds an unnecessary degree of freedom to the experiment. The only approach that would allow us to control the distance between expert estimates and true values (and, thus, neutralize the impact of human factor) is simulation of expert estimates themselves. That is why it is relevant to use the simulation-based approach [15] for comparison of methods in terms of efficiency.

## 6 Comparing ordinary and modified combinatorial methods through simulation of expert estimation of model objects

The key idea of data aggregation methods' efficiency evaluation is verification of their stability under fluctuations of input data (i.e. PCM). As we mentioned in the introduction, it is assumed that there is a certain true value of the object's estimate according to a given criterion (such as exact number of colored pixels in a picture). The estimate provided by an expert differs from this true value by estimation error. Let us assume that under the same expert estimation errors one aggregation method produces the result (priority vector), that is closer to the model vector (of true values) than the result produced by the other method. In this case we can state that the first method is more efficient. In order to be able to monitor expert estimation errors and deviations of priority vectors, obtained by different methods, from model values (in the context of aggregation method comparison), let us simulate expert estimation process in the following way.

1. Set true model object weights.
2. Built an ICPCM $A$ (based on the rule $a_{ij} = w_i / w_j$ for multiplicative or $a_{ij} = w_i - w_j$ for additive comparisons, where $a_{ij}$ is the element of $A$).

3. After that, add a certain "noise" to matrix $A$ so that each element (except diagonal ones) changes as follows: $a'_{ij} = a_{ij} \pm a_{ij} \cdot \delta / 100\%$, where $\delta > 0$ is a value set in advance, which defines maximum relative deviation of pair-wise comparisons provided by the expert (i.e. elements of $A$) as percentage of true values. In this way expert estimation errors are simulated. In this case $\delta$ denotes potential relative error made by the expert during pair-wise comparison session.

4. "Perturbed" PCM $A'$ is used as input data for one of aggregation methods, which produces aggregate object weights $w'_i$ (priority vector). We propose to define the efficiency of expert data aggregation method based on maximum possible relative deviation of calculated object weight from true value of the same weight. The method producing smaller deviations should be considered more efficient.

$$\Delta = \max_i \left| \frac{w'_i - w_i}{w_i} \right| \times 100\% \quad (9)$$

Calculated values of $\Delta$ will depend on both $\delta$ and true relative weight values, set by experiment organizer. That is why the values of efficiency indicator for each method are presented in the form of function $\Delta(\delta)$ for each priority vector variant.

Dependence $\Delta(\delta)$ is defined for each method on an interval $\delta \in \,]0;100[$ (we assume that relative pair-wise comparison error made by an expert should not exceed 100%, although in the general case the function $\Delta(\delta)$ is defined in a wider range $\delta \in \,]0;\infty[$). We propose to find maximum possible deviation $\Delta$ for every $\delta$ using the genetic algorithm (GA) [21].

Just like in the previous section, we used an original software module to conduct the experiment. The module allows us to generate and perturb the initial PCM and launch the GA (Fig. 5).

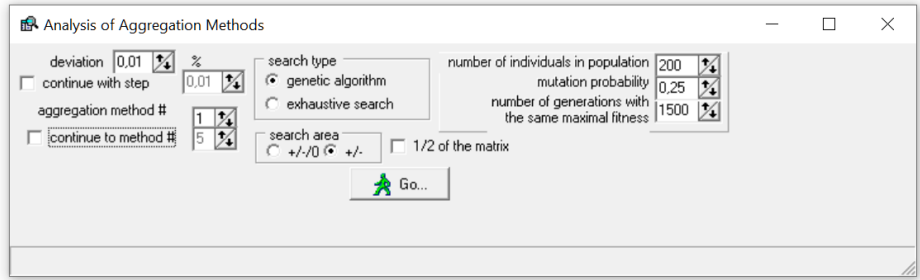

**Fig. 5**. Screenshot of the software module for experimental research of combinatorial method using the GA

In terms of the GA, the "individuals" are the perturbed PCM with given relative error value $\delta$. "Fitness function" is the maximum relative deviation of resulting priorities from the true value of object weights $\Delta$ (9). The GA works as follows.

1. For given true values of object weights and $\delta$ a "population" of individuals (perturbed PCM) is generated. (Cardinality of complete enumeration of matrices of dimensionality $n \times n$, whose elements differ from true values by $\delta$, is very large, so a population includes only a part of PCM from the complete set. However, for $n \leq 5$ complete enumeration is possible).

2. Individuals with maximum fitness function value are selected from the population. That is, we are selecting PCM, which produce maximum deviation of priority vector from true value after aggregation.

3. Individuals from the selected subset are "interbred" through weighted summation (convex combination) and mutation. As a result, we get a new generation of individuals.

4. If for the new generation the value of fitness function $\Delta$ is larger than for the previous one, we should move to step 2. If during a fixed number of generations $\Delta$ does not grow, then the algorithm stops and terminates its work. As a result we get maximum $\Delta$ for a given $\delta$.

In essence, we are looking for a maximum of a function of many variables $f(a'_{ij}); i,j = (1,n)$. Its arguments are elements of PCM $A'$. Values of $\Delta(\delta)$ for each aggregation method also depend on specific values of initially set true values of object weights $w_i, i = (1,n)$. Examples of functions $\Delta(\delta)$ for given true model values of object weights are shown on Fig. 6. Variants of model weight vectors are set in a way that illustrates different ratios between object weights (equal, equal in pairs, arithmetic progression, geometric progression, extreme values of scale range etc).

In [15] individual "weighted" combinatorial method was compared with several other individual pair-wise comparison aggregation methods (Fig. 6). In the context of the present paper we should note that row geometric mean (one of the methods, studied in [15]) is equivalent to ordinary combinatorial method (as proven in [19]). So, the advantage of modified (weighted) combinatorial method over row geometric mean entails its advantage over the ordinary method ([15, 18]) and LLSM (that is also equivalent to ordinary method [8]).

Consequently, we can conclude that empirical comparison data of several modifications of pair-wise comparison aggregation methods (obtained through simulation of expert estimation of model object weights) confirm the advantage of modified combinatorial method over other methods. The efficiency indicator is its stability to perturbations of the initial PCM (that is, to expert's errors).

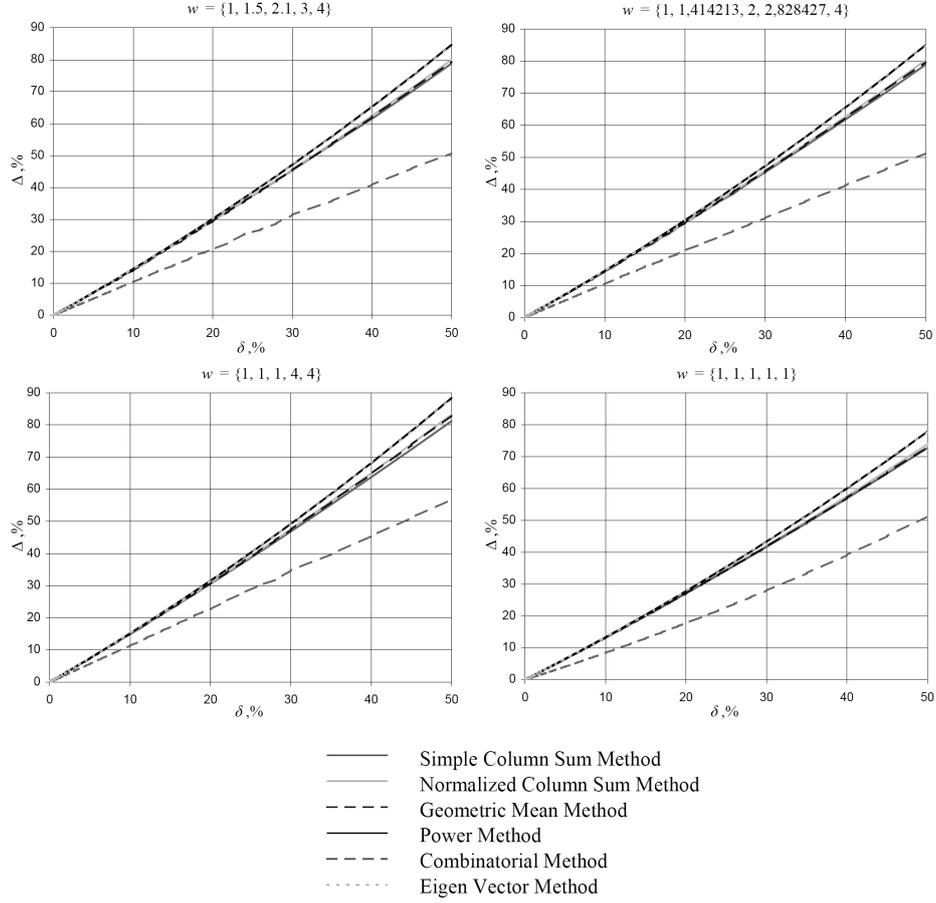

**Fig. 6**. Examples of dependence of $\Delta$ on $\delta$ for different model priority vectors

We should note that experimental results, shown on Fig 6, are obtained only fro individual estimation methods ($m=1$), mostly, in additive scale. That is, in the conducted experiments in formulas (1) and (2) product is replaced by summation.

$$w_j^{aggregate} = (1/T)\sum_{q=1}^{T} w_j^q, j=1..n, T \in [1..n^{n-2}] \quad (10)$$

$$w_j^{aggregate} = (1/\sum_{u,p,v} R_{upv})\sum_{k,l=1}^{m}(\sum_{q^k=1}^{T_k} R_{kq_k l} w_j^{(kq_k l)}), j=1..n \quad (11)$$

The next phase of the research will include simulation of group expert estimate aggregation process, where the estimates are provided in both additive and multiplica-tive scales. Transition from multiplicative to additive scales (particularly, during modeling) can be performed using logarithm and power operators. Both functions are monotonously increasing, so the properties of additive and multiplicative methods should be similar.

## 7      Conclusions

We have considered two possible ways of evaluating the efficiency of pair-wise comparison aggregation methods, namely, combinatorial methods of pair-wise comparison aggregation, where spanning tree weights are taken and not taken into account. We have shown that traditional concept of accuracy does not apply to expert data-based methods. As a result, simulation turns out to be the more correct way of evaluating the efficiency of the methods, than their testing on real expert data. We have suggested two approaches to efficiency evaluation of combinatorial method of pair-wise comparison aggregation (individual and group, additive and multiplicative methods): 1) defining relative accuracy of real expert pair-wise comparison aggregation results and 2) simulation of the whole expert examination lifecycle. We have obtained experimental results, that empirically prove the advantage of the modified combinatorial method over the ordinary method (and, consequently, over row geometric mean and LLSM).

Experimental results allow us to draw some fundamental conclusions: 1) the concepts of accuracy of expert estimates and efficiency of expert data aggregation methods should be clearly distinguished; low accuracy of some method's results is often induced by experts' errors, and not by drawbacks of the method itself; 2) it is not the accuracy, but "consistent accuracy" that counts, both within a PCM of 1 expert and in a group of experts; more consistent and compatible estimation results should be considered more credible; 3) real expert data can be used to compare conceptually different methods, using expert information of different types; if input information for several methods is the same, and only aggregation procedures are different, then in order to evaluate and compare such methods, we should simulate the whole expert session cycle, including the estimates themselves. Further research will be targeted at extended studies of modified combinatorial method through simulation of group expert estimates' aggregation for the cases when objects are estimated in both additive and multiplicative scales.